\documentclass[conference]{IEEEtran}
\IEEEoverridecommandlockouts
\usepackage{bm}
\usepackage{cite}
\usepackage{amsmath,amssymb,amsfonts}
\usepackage{algorithmic}
\usepackage[lined,ruled,boxed,linesnumbered]{algorithm2e} 
\usepackage{textcomp}
\usepackage{xcolor}
\usepackage{subfigure,epsfig,color,longtable}
\usepackage{graphicx,psfrag,url,stfloats,amsmath,multirow}

\def\BibTeX{{\rm B\kern-.05em{\sc i\kern-.025em b}\kern-.08em
    T\kern-.1667em\lower.7ex\hbox{E}\kern-.125emX}}
    
\makeatletter
\DeclareRobustCommand\onedot{\futurelet\@let@token\@onedot}%
\def\@onedot{\ifx\@let@token.\else.\null\fi\xspace}%

\makeatother

\makeatletter

\newcommand{\Rmnum}[1]{\expandafter\@slowromancap\romannumeral #1@}

\makeatother
\begin{document}

\title{Context-guided Triple Matching for Multiple Choice Question Answering\\
\thanks{ }
}

\author{
\IEEEauthorblockN{Xun Yao}
\IEEEauthorblockA{
\textit{\small{Wuhan Textile University, China}}\\
\small{yaoxun@wtu.edu.cn}}
\and
\IEEEauthorblockN{Junlong Ma}
\IEEEauthorblockA{
\textit{\small{Wuhan Textile University, China}}\\
\small{1137436221@qq.com}}
\and
\IEEEauthorblockN{Xinrong Hu}
\IEEEauthorblockA{
\textit{\small{Wuhan Textile University, China}}\\
\small{hxr@wtu.edu.cn}}
\and
\IEEEauthorblockN{Junping Liu}
\IEEEauthorblockA{
\textit{\small{Wuhan Textile University, China}}\\
\small{jpliu@wtu.edu.cn}
}
\and
\IEEEauthorblockN{Jie Yang}
\IEEEauthorblockA{
\textit{\small{University of Wollongong, Australia} }\\
\small{jiey@uow.edu.au}}
\and
\IEEEauthorblockN{Wanqing Li}
\IEEEauthorblockA{
\textit{\small{University of Wollongong, Australia} }\\
\small{wanqing@uow.edu.au}}
}

\maketitle

\begin{abstract}
The task of multiple choice question answering (MCQA) refers to identifying a suitable answer from multiple candidates, by estimating the matching score among the \emph{triple} of the passage, question and answer. Despite the general research interest in this regard, existing methods decouple the process into several pair-wise or \emph{dual} matching steps, that limited the ability of assessing cases with multiple evidence sentences.
To alleviate this issue, this paper introduces a novel \textbf{C}ontext-guided \textbf{T}riple \textbf{M}atching algorithm, which is achieved by integrating a Triple Matching (TM) module and a Contrastive Regularization (CR). The former is designed to enumerate one component from the triple as the background context, and estimate its semantic matching with the other two. Additionally, the contrastive term is further proposed to capture the dissimilarity between the correct answer and distractive ones.
We validate the proposed algorithm on several benchmarking MCQA datasets, which exhibits competitive performances against state-of-the-arts.
\end{abstract}

\section{Introduction}
Question answering turns out to be one of the most popular and challenging research topics in machine reading comprehension (MRC). Existing studies of question answering focus on either discovering (\emph{extracting}) spans from the given passage \cite{seonwoo-etal-2020-context, SpanBERT}, or identifying (\emph{selecting}) the most suitable answers to questions from a set of candidates, known as multiple choice question answering (MCQA) \cite{Duan2021, LI2021106936, Zhang2020}. This paper is on a novel method for MCQA.

Approaches to MCQA usually consist of a two-step process. In the first step, words in the passages, questions and candidate answers are encoded, using a pre-trained language model, into fixed length of vectors. The second step is to generate a representation by exploring the semantic-relationship matching among a passage, question, and answer.

In general, improvement of methods for MCQA can be achieved by fine-tuning the pre-trained model to the context of the passage, question and answer and/or improving the subsequent matching \cite{zhang2019sgnet, zhu2020duma}. Typical work on the former include those in \cite{XLNet, megatronlm, Yinhan2019}. A recent work on the latter is \textit{DCMN+}\cite{Zhang2020}, where conventional unidirectional matching is extended to bidirectional matching among the pairs of question-passage, question-answer and passage-answer. 
The bidirectional matching improves the capability of capturing the semantic relationship among the triple (\emph{i.e}. passage, question and answer), hence, the performance compared with uni-directional matching. However, such pairwise matching, though bi-directional, ignores the knowledge from one entity of the triplet, which has limited its ability to deal with cases where there are multiple evidence sentences in the passage with respect to the question and answer. Table~\ref{tab:motivating} shows an example from the popular MCQA dataset (RACE \cite{race}), where answering this cloze question depends on multiple evidence sentences in the passage as highlighted in the table.

\begin{table}[!th]
\centering
\caption{An example from the RACE dataset \cite{race}. The evidence sentences (from the passage) and keywords (from the question and candidate answers) are highlighted. DCMN+ \cite{Zhang2020} picks answer A where the correct answer is B.}
\begin{tabular}{|p{0.47\textwidth}|}
\hline
\textit{Question}: According to the passage, when we become \textbf{adults}, \_\_\_\_\_?   \\ \hline
\textit{Passage}: Most people believe they don't have imagination. $\cdots$ but most of us, \textbf{once we became adults}, \textbf{forget how to access it}. Creativity isn't always connected with great works of art or ideas. \textbf{People at work and in their free time} \textbf{routinely think of creative ways to solve problems}. $\cdots$ \textbf{Here are three techniques to help you.} $\cdots$ \\ \hline
\textit{Answers}:\\
$\textit{A}.$ most of us are no longer \textbf{creative}; \\
$\textit{B}.$ we can still learn to be more \textbf{creative}; \\
$\textit{C}.$ we are not as imaginative as children; \\
$\textit{D}.$ we are unwilling to be \textbf{creative};
 \\ \hline
\end{tabular}
\label{tab:motivating}
\end{table}

To address this issue, this paper proposes to extend the DCMN+ \cite{Zhang2020} to \textbf{C}ontext-guided \textbf{T}riple \textbf{M}atching algorithm (CTM). Specifically, a context is provided via the missing entity in performing the conventional pair-wise matching. In other words, CTM performs matching with respect to a context (an entity from the triple) to exploit the semantic relationship more specifically than DCMN+. In addition, a contrastive regularization (CR) is adopted in strengthening the learning of the semantic differences among answer candidates. This regularization follows a recently proposed self-supervised learning diagram, \emph{i.e}., \textit{contrastive learning}, which helps in differentiating keywords in the candidate answers (such as ``\textbf{creative}") as illustrated Table~\ref{tab:motivating}.

The contributions of the paper include:
\begin{itemize} \setlength{\itemsep}{5pt}
	\item context is introduced into the matching process, and a context-guided triplet matching is proposed accordingly in order to improve the ability in effectively capturing semantic relationship from a passage, questions and answers; and
	\item contrastive regularization is developed to learn distinctive features among similar candidate answers; and
	\item extensive experiments are conducted on two widely used MCQA datasets to evaluate the proposed CTM, and state-of-the-art results are achieved in comparison with the existing methods. 
\end{itemize}

Our code will be publicly available from Github. 

\section{Related work}\label{sec:background}
In this section, we provide background information on the study area, focusing on existing work on MCQA and the concept of contrastive learning. 

\subsection{MCQA}
Multiple choice question answering (MCQA) is a long-standing research problem from machine reading comprehension, where the key is to determine one correct answer (from all candidates) given the background passage and question. Several models have been proposed which utilize deep neural networks with different \textbf{matching} strategies.

Chaturvedi \emph{et al}. first concatenate the question and candidate answer, and calculate the matching degree against the passage via attention \cite{chaturvedi-etal-2018-cnn}.
The work \cite{wang2018} treats the question and a candidate answer as two sequences before matching them individually with the given passage. Then a hierarchical aggregation structure is constructed to fuse the previous co-matching representation to predict answers.
Similarly, a hierarchical attention flow is proposed in \cite{Zhu2018} to estimate the matching relationship based on the attention mechanism at different hierarchical levels.
Zhang \emph{et al}. propose a dual co-matching network in \cite{Zhang2020}, which formulates the matching model among background passages, questions, and answers bi-bidirectionally. 

Apart from the aforementioned matching-based work, another line of studies proposes to integrate with the auxiliary knowledge. For instance, a syntax-enhanced network is presented in \cite{zhang2019sgnet} to combine syntactic tree information with the pre-trained encoder for better linguistic matching.
Duan \emph{et al}. utilize the semantic role labeling to enhance the contextual representation before modeling the correlation \cite{Duan2021}. More recently, the off-the-shelf knowledge graph is leveraged to fine-tune the downstream MCQA task in \cite{LI2021106936}.

Compared to existing matching work in \cite{wang2018, Zhang2020}, the proposed algorithm performs matching by introducing a context (an entity from the triple of passage, question and answer). This context serves as a background knowledge to exploit the semantic relationship with the remaining two entities.

\subsection{Contrastive learning}\label{sec:cl}
Contrastive learning (CL) has attracted a lot of research attention in the last several years, which has prohibited promising results in many downstream tasks, such as text clustering \cite{gao2021simcse},machine translation \cite{liang2021rdrop}, and knowledge graph completion \cite{Qin2020}, and \emph{etc}.

The main idea is to leveraging input data itself for self-supervised training. In particular, for a given anchor sample $\bm{x}_i$, one encoder $f(\cdot)$, and a pre-defined similarity function $sim()$, CL aims to optimize the following loss function:
\begin{equation}\label{eq:traditional_CL}
	sim\left( f(\bm{x}_i), f(\bm{x}^+_i) \right) \gg sim\left( f(\bm{x}_i), f(\bm{x}^-_i) \right),
\end{equation}
where $\bm{x}^+_i$ and $\bm{x}^-_i$ are contrastive samples of $\bm{x}_i$. The subsequent training is to assign large values to positive samples $\bm{x}^+_i$ and small values to the negative ones $\bm{x}^-_i$. In this paper, our motivation is to integrate CL with MCQA to capture and enhance the representation difference between the correct and distractive answers, that has not been explored before.

\section{Proposed method}\label{sec:proposedmethod}
\begin{figure}[!tph]
	\centering
	\includegraphics[width=0.478\textwidth]{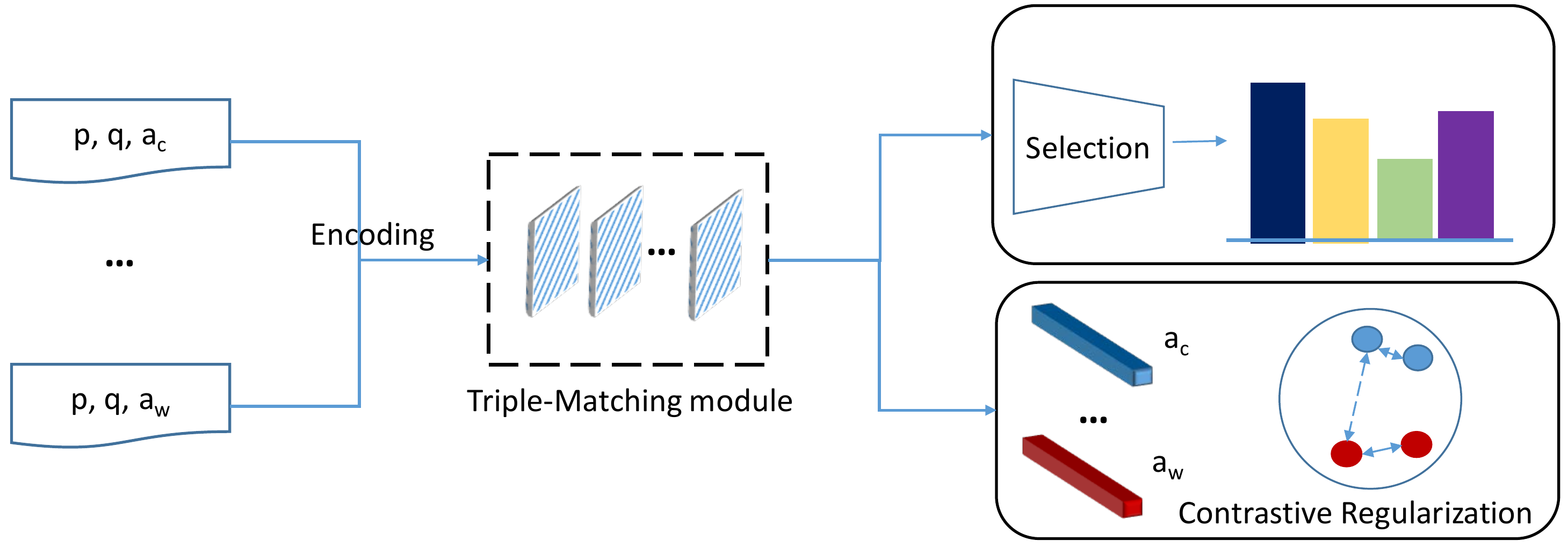} 
	\caption{Overview of the proposed \textbf{C}ontext-guided \textbf{T}riple \textbf{M}atching algorithm for MCQA ($p$-passage, $q$-Question, $a_c$-correct answer, $a_w$-wrong answer). }
	\label{fig:overall}
\end{figure}

The proposed method gradually identifies the best-matching answer by coordinating the loss from a Triple Matching (TM) module and a Contrastive Regularization (CR) simultaneously, as illustrated in Fig. \ref{fig:overall}.

Given an input triple of passage, question and answer, a pre-trained language model is first utilized for \textit{encoding} textual contents. Then the \textit{Triple}-\textit{Matching} module enumerates this input triple and selects one entity as the background context. The semantic relationship is accordingly estimated using the remaining two entities with regard to this selected context. At last, the produced features from TM are utilized for answer \textit{selection}, while the \textit{contrastive regularization} ensures that the feature agreement between correct answers is maximized, by contrasting to the agreement between distractive ones.

\subsection{Encoding}

Let $p$, $q$ and $a$ be a passage, a question and a candidate answer, respectively. A pre-trained model (\emph{e.g}. $BERT$) is adopted to encode each word from them into a fixed-length vector, yielding
\begin{equation}
	\bm{H}^p = Enc(p),\bm{H}^q = Enc(q),\bm{H}^a = Enc(a),
\end{equation}
where $\bm{H}^p\in\mathbb{R}^{|p| \times l }$,$\bm{H}^q\in\mathbb{R}^{|q| \times l }$,and $\bm{H}^a\in\mathbb{R}^{|a| \times l }$ are relevant representation of $p$, $q$, and $a$, respectively, and $l$ is the dimension of the hidden state.

\subsection{Triple matching}

To model the relationship among the triple of \{$p$, $q$, $a$\}, in TM we introduce an context-oriented mechanism. That is, we select one component from the triple once (as the background context), and estimate its semantic correlation with the remaining two to further produce a context-guided representation. Note that this proposed module involve all three entities from the triple simultaneously, while existing methods adopt the pairwise strategy that involves only two entities once.

Taking the answer $a$ as an example, below we show how to model the representation for the answer(context)-guided passage-question matching.
At first, given the encoder output of $\bm{H}^p$, $\bm{H}^a$ and $\bm{H}^q$, we apply the bidirectional attention to calculate the answer-aware passage representation ($\bm{E}^p \in \mathbb{R} ^ {|a| \times l}$) and answer-aware question representation ($\bm{E}^q \in \mathbb{R} ^ {|a|\times l}$) as follows:
\begin{equation}
	\begin{split}
		\bm{G}^{aq}  &= SoftMax(\bm{H}^aW\bm{H}^{qT}) \\
		\bm{G}^{ap}  &= SoftMax(\bm{H}^aW\bm{H}^{pT}) \\
		\bm{E}^p     &= \bm{G}^{ap}\bm{H}^p,\bm{E}^q = \bm{G}^{aq}\bm{H}^q,
	\end{split}
	\label{eq:att_pq}
\end{equation}
where $W \in \mathbb{R}^{l \times l}$ are learnable parameters, and $\bm{G}^{aq} \in \mathbb{R}^{|a| \times |q|} $ and $\bm{G}^{ap} \in \mathbb{R}^{|a| \times |p|} $ are the attention matrix between the answer-question, and the answer-passage, respectively. 

Next, we further allow the third entity to be included by adopting the bidirectional attention again (to embed the question for $\bm{E}^p$ and the passage for $\bm{E}^q$). As a result, the core of triple matching becomes:
\begin{equation}
	\begin{split}
		\bm{G}^{pq}  &= SoftMax(\bm{E}^pW_1\bm{E}^{qT}) \\
		\bm{G}^{qp}  &= SoftMax(\bm{E}^qW_1\bm{E}^{pT}) \\
		\bm{E}^{pqa} &= \bm{G}^{pq}\bm{H}^a, \bm{E}^{qpa} = \bm{G}^{qp}\bm{H}^a \\
		\bm{S}^{pqa} &= ReLU(\bm{E}^{pqa}W_2) \\
		\bm{S}^{qpa} &= ReLU(\bm{E}^{qpa}W_2),
	\end{split}\label{eq:bef_att_pqa}
\end{equation}
where $W_1$, $W_2$ $\in\mathbb{R}^{l \times l}$ are learnable parameters, and $\bm{E}^{pqa} \in \mathbb{R} ^ {|a|\times l}$, $\bm{E}^{qpa} \in \mathbb{R} ^ {|a|\times l}$ represent passage-question-aware answer representation and question-passage-aware answer representation, respectively. The final representation of answer-guided passage-question matching (\emph{i.e}. $\bm{M}^{\underline{a}}\in \mathbb{R}^{2 \times l}$) is to aggregate the above as follows:

\begin{equation}
	\begin{split}
		\bm{M}^{pqa} &= MaxPooling(\bm{S}^{pqa}) \\
		\bm{M}^{qpa} &= MaxPooling(\bm{S}^{qpa}) \\
		\bm{M}^{\underline{a}}     &= [\bm{M}^{pqa}; \bm{M}^{qpa}], 
	\end{split}
	\label{eq:att_pqa}
\end{equation}
where the aggregated representation $\bm{M}^{pqa} \in \mathbb{R}^l$ and $\bm{M}^{qpa} \in \mathbb{R}^l$ is computed via a row-wise max pooling operation from Eq. (\ref{eq:bef_att_pqa}).

\begin{figure*}[!tph]
	\centering
	\includegraphics[width=0.78\textwidth]{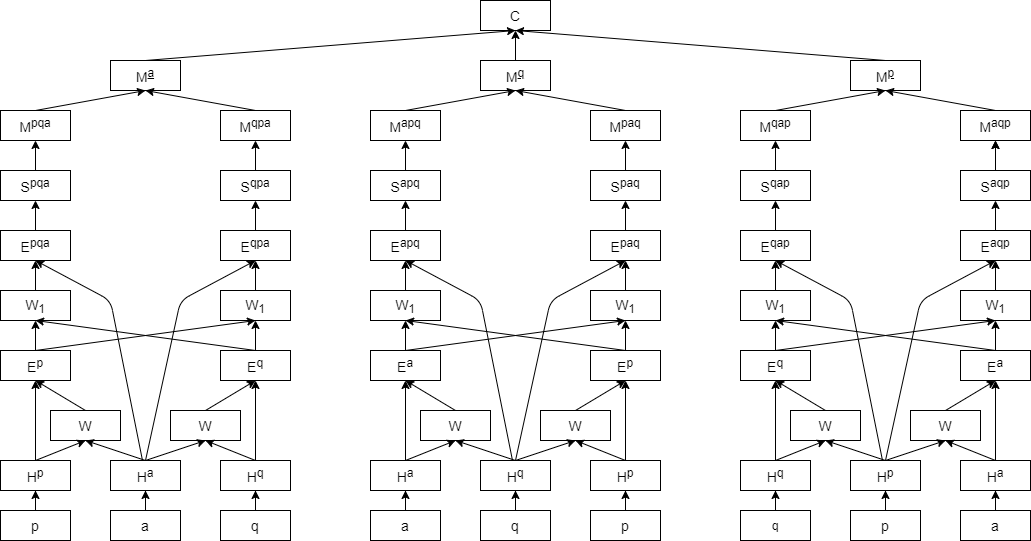} 
	\caption{Proposed Triple Matching (TM) module that consists of three components, including the representation of answer-guided passage-question matching ($\bm{M}^{\underline{a}}$), question-guided passage-answer matching ($\bm{M}^{\underline{q}}$), and passage-guided answer-question matching ($\bm{M}^{\underline{p}}$), respectively. For each component, we aim to estimate the semantic correlation between one selected context and anther two entities. }
	\label{fig:tm}
\end{figure*}

Similarly, we enumerate the other two entities (that is, the question $q$ and passage $p$) to compute the related representation for the question-guided answer-passage matching (\emph{ie}., $\bm{M}^{\underline{q}} \in \mathbb{R}^{2 \times l}$) and the passage-guided answer-question matching (\emph{ie}., $\bm{M}^{\underline{p}} \in \mathbb{R}^{2 \times l}$), following the same procedure from Eq.(\ref{eq:att_pq}) to Eq.(\ref{eq:att_pqa}). In sum, the proposed TM module is illustrated in Figure \ref{fig:tm}.

\subsection{Answer selection}

With the triple-matching representations $\bm{M}^{\underline{a}}$, $\bm{M}^{\underline{q}}$, $\bm{M}^{\underline{p}}$, we further concatenate them as the final representation $\bm{C}$ (\emph{ie}., $\bm{C} = [\bm{M}^{\underline{a}}; \bm{M}^{\underline{q}}; \bm{M}^{\underline{p}}]$). Let $\bm{C}_i$ be the representation for the $\{p, q, a_i\}$ triplet. Accordingly, the selection loss can be computed as follows:
\begin{equation}\label{eq:tm}
		\mathcal{L}_{TM}(p, q, a_i) = -log\frac{\exp(\bm{C}_iV)}{\sum_{i=1}^{n}\exp(\bm{C}_iV)}
\end{equation}
where $V \in \mathbb{R}^l$ is a learnable parameter and $n$ is the number of answer options.

\subsection{Contrastive regularization}
The aforementioned TM module is performed to extract semantic representation from one candidate triple. Yet, there could be trivial (word) difference between the correct and distractive answers (see Table \ref{tab:motivating}). To highlight this dissimilarity, we accordingly propose to utilize a contrastive regularization. The purpose is to maximize the agreement from correct answer(s) via pushing away the agreement against distractive ones.

To apply we first need to form contrastive (\emph{i.e}. both \textit{positive} and \textit{negative}) samples for an anchor. MCQA is enjoyed owing to those distractive answers, which in nature play a role of negative samples against the correct answer. For positive ones, we adopt a similar strategy by following the dropout operation from \cite{gao2021simcse, liang2021rdrop}.
More precisely, given the correct triple of $\{p, q, a_c\}$, we simply apply the TM module \textit{twice} with different dropout masks to produce the anchor and positive representation as $\bm{C}_c$ and $\bm{C}^{+}_c$, respectively.
Then the contrastive-regularized learning objective can be defined as follows:
\begin{equation}
	\begin{split}
		\mathcal{L}_{CR}(p, q, a_c) &= -log\frac{\exp^{cos(\bm{C}_c, \bm{C}^+_c)/ \tau}}{\sum_{i=1}^{n}\exp^{cos(\bm{C}_c, \bm{C}_i)/ \tau}},
	\end{split}
\end{equation}
where $n$ is the size of the mini-batch including one anchor, positive and negative samples, and $\tau$ is a pre-defined temperature.

\subsection{Loss function}
With two losses from the answer selection and contrastive regularization, we propose to train the model using the \textit{joint} loss as follows:
\begin{equation}\label{eq:final}
\mathcal{L} = \mathcal{L}_{TM} + \lambda_{CR} \mathcal{L}_{CR},
\end{equation}
where $\lambda_{CR}$ is a penalty term\footnote{
There are another two training strategies, including \textit{pre-train} and \textit{alternate}. The former is to update the model first using $\mathcal{L}_{CR}$ before finetuning with $\mathcal{L}_{TM}$, while the latter is to train the model with $\mathcal{L}_{TM}$ for $(N_t-1)$ iterations and switch to $\mathcal{L}_{CR}$ once, for every $N_t$ iterations. However, the experimental results show the \textit{joint} training outperforms \textit{pre-train} and \textit{alternate} based model.
}.

\subsection{Discussion}\label{sec:connection}
Next, we analyze the relationship between the proposed method and existing pairwise algorithms. Previous studies measure the matching representation (\emph{i.e}. $\bm{C}$ from Eq. (\ref{eq:tm})) using the following estimation:
\begin{itemize} \setlength{\itemsep}{5pt}
\item CNN-Matching \cite{chaturvedi-etal-2018-cnn}: 
\begin{equation}
	\begin{aligned}
		\bm{H}^{qa} = &Enc([q;a]); \bm{H}^{p} = Enc(p); \\
		\bm{M} = &Att(\bm{H}^{qa},\bm{H}^{p}); \\
		\bm{C} = &Sim(\bm{H}^{qa}, \bm{M}). \\
	\end{aligned}
\end{equation}	

\item Co-Matching \cite{wang2018}: 
\begin{equation}
	\begin{aligned}
		\bm{H}^q = &Enc(q); \bm{H}^a = Enc(a); \bm{H}^p = Enc(p); \\
		\bm{M}^{qp} = &Att(\bm{H}^q, \bm{H}^p); \bm{M}^{ap} = Att(\bm{H}^a, \bm{H}^p);  \\
		\bm{C} = &[Sim(\bm{M}^{qp}, \bm{H}^p); Sim(\bm{M}^{ap}, \bm{H}^p)].\\
	\end{aligned}
\end{equation}	

\item DCMN+ \cite{Zhang2020}: 
\begin{equation}
	\begin{aligned}
		\bm{H}^q = &Enc(q); \bm{H}^a = Enc(a); \bm{H}^p = Enc(p); \\
		\bm{M}^{qa} = &Att(\bm{H}^q, \bm{H}^a); \bm{M}^{qp} = Att(\bm{H}^q, \bm{H}^p); \\
		\bm{M}^{ap} = &Att(\bm{H}^a, \bm{H}^p); \\
	   \bm{C} =	& [ Gat(\bm{M}^{qa}, \bm{M}^{ap}) ; Gat(\bm{M}^{qp}, \bm{M}^{ap});\\
		 			& Gat(\bm{M}^{qa}, \bm{M}^{qp}) ].\\
		\end{aligned}		
\end{equation}	
\end{itemize}
Within aforementioned methods, $Enc$ represents the encoder, $Att$ stands for the attention operation, $Sim$ is for the similarity calculation, $Gat$ is a rest gate function, and $[;]$ is the vector concatenation. Note that existing methods adopted different implementation of $Enc$, $Att$, and $Sim$, \emph{etc}. For instance, $Enc$ in \cite{chaturvedi-etal-2018-cnn} and \cite{wang2018} has been implemented as CNN and BERT, respectively. 

Compared to the aforementioned methods, the proposed algorithm can be cast as their extension, with an additional consideration of triple matching and contrastively representing the correct answer(s). 
That is, the triple matching is to apply two attention layers to estimate the semantic relationship with regard to the selected context. As such, Eq.(\ref{eq:att_pq}) to Eq.(\ref{eq:att_pqa}) can be equivalently represented as the following process:
\begin{equation}
	\begin{aligned}
		\bm{M}^{qa} = &Att(\bm{H}^q, \bm{H}^a); \bm{M}^{pa} = Att(\bm{H}^p, \bm{H}^a);  \\		
		\bm{M}^{pqa} = &Att( Att(\bm{M}^{pa}, \bm{M}^{qa}),  \bm{H}^a); \\
		\bm{M}^{qpa} = &Att( Att(\bm{M}^{qa}, \bm{M}^{pa}),  \bm{H}^a).
		\end{aligned}		
\end{equation}	
%
In addition, our method is also distinct from existing ones by further integrating the contrastive loss. That is, we aim to distinguish the correct answers via pulling its relevant representation away from distractive ones, which has been neglected by existing pairwise-matching approaches.

\section{Experiments}\label{sec:exp}
The proposed CTM method is evaluated on two widely used MCQA dataset and compared to the state-of-the-art methods.

\subsection{Datasets}

The two datasets adopted in the experiments are  RACE \cite{race} and DREAM \cite{sundream2018}. RACE is one of the widely used banchemark datasets developing and evaluating methods for multi-choice reading comprehension. It consists of subsets RACE-M and RACE-H that correspond to the reading-difficulty level of middle and high school, respectively.

DREAM is a dialogue-based examination dataset. It includes dialog passages as the background and three options associated with each individual question.

Table \ref{tab:dataset} shows the statistics of the two datasets, including total number of passages, number of questions, average number iof candidate answers and average number of words per candidate answer. In particular, we notice that the averaged length per answer over the three datasets, RACE-M, RACE-H and DREAM, is approximately 5.7 words.

\begin{table}[!th]
{\small
\caption{Summary of RACE and DREAM, where \textbf{\#a} is the averaged number of candidate answers per question, and \textbf{\#w/a} is the averaged length per answer.}\label{tab:dataset}
\begin{center}\begin{tabular}{|c|c|c|c|c|}
\hline
\textbf{Dataset}	& \textbf{passages} & \textbf{Questions} & \textbf{\#a} & \textbf{\#w/a}\\ \hline
RACE-M   &	7,139   &   28,293   &   4 & 4.9\\ \hline
RACE-H   &	20,794   &   69,394   &   4 & 6.8\\ \hline
DREAM    &   6,444    &  10,197    &  3 & 5.3 \\ \hline
\end{tabular}
\end{center}
}
\end{table}

\subsection{Implementation and settings}
Two pre-trained language models, including the BERT$_{base}$ and BERT$_{large}$, are adopted as the encoder for word-embedding.  
BERT$_{base}$ consists of 12-layer transformer blocks, 12 self-attention heads, and 768 hidden-size, whereas BERT$_{large}$ consists of 24-layer transformer blocks, 16 self-attention heads, and 1024 hidden-size. They have 110M and 340M parameters, respectively. The dropout rate for each BERT layer is set as 0.1. The Adam optimizer with a learning rate setting of $2e^{-5}$ is adopted to train the proposed CTM. 

During training, batch size is 4, number of training epoch is 3, and the max length of input sequences is set to 360 for RACE. For DREAM, batch size is 4 and number of training epochs is 6, and the max length of input sequences is set to 300. With passages of more than 360/300 words in RACE or DREAM, we follow the same sliding window strategy as that in \cite{jin2019mmm} to split the long passage into sub-passages of length 360/300 allowing overlapping content.

For the contrastive regularization, the dropout rate is 0.1 to produce positive samples, and the temperature $\tau = 0.07$.
The CTM model is trained on a machine with four Tesla K80 GPUs. 
Accuracy $acc = n_q^+/n_q$ is used to measure the performance, where $n_q^+$ represents the number of questions that the model selects the correct answer, and $n_q$ is the number of total questions.

\subsection{Results}\label{sec:comparison}
We comared the performance of the proposed CTM with the methods, including the public models from the leaderboard (\emph{i.e}. BERT) and state-of-the-arts (\emph{i.e}. DCMN+\cite{Zhang2020}). To make a fair comparison, we are particularly interested in those implemented with the same BERT encoder. Yet, some public works (such as MegatronBERT \cite{megatronlm} and RoBerta \cite{Yinhan2019}) may produce better performance. The improvement is likely because they use much larger model size (\emph{e.g.}, 3.9 billion parameters for MegatronBERT) or complex pre-training strategies (for RoBerta). As such the results are not strictly comparable to the proposed CTM method.\footnote{Given the availability of numerous pre-training models, we could simply replace the adopted BERTs with other more powerful encoders, such as \cite{Yinhan2019}, to improve the performance of CTM. Alternatively, to use the additional ground knowledge, such as the work \cite{zhang2019sgnet}, could also lead to a potential improvement for CTM. We leave these as the future work.}

Results of the proposed CTM and comparing methods are shown in Table \ref{tab:main_res}. It can be seen that the proposed method achieves state-of-the-art performance on both RACE and DREAM datasets. Not surprisingly, the BERT$_{base}$ method and methods using BERT$_{base}$  achieve generally worse performance compared to the counterparts using BERT$_{large}$, which shows the contribution that a good pre-trained model would bring. 

Although baseline performance (from BERTs) is further improved by using bidirectional matching~\cite{Zhang2020} or external knowledge \cite{zhang2019sgnet}, these strategies do the pairwise among the passage, question and answer independently without considering the third entity in the triplet. By contrast, the proposed CTM method uses the third entity to provide a background context during the matching, so that learned features are geared towards this selected context. The use of contrastive regularization further strengthens the learning to differentiate the correct answer from semantically closed but wrong ones. As a result, the proposed CTM substantially outperforms these state-of-the-art methods.

\begin{table*}[!htbp]
\caption{Results in accuracy (\%),obtained by CTM and the comparing methods on the on the test set. 
In the table, ``-B'' and ``-L'' represents the \textit{base} and \textit{large} model of the BERT encoder.
``$\times$'' indicates there is no results from the original reference, and ``$\bigstar$'' shows the original reference doesn't differentiate RACE-M and RACE-H but only report the averaged result.}\label{tab:main_res}
\begin{center}
\begin{tabular}{l|ccc}
\hline
\textbf{Algorithm}	& \textbf{RACE-M} 		& \textbf{RACE-H} 		& \textbf{DREAM} 			\\ \hline
BERT(-B)											&	71.1   					& 62.3     					&   63.2			\\ 
BERT(-L)											&	76.6   					& 70.1 						&   66.8			\\ 
Human Ceiling 									&  95.4						& 94.2						& 	98.6		\\ \hline
\hline

DCMN+(-B)\cite{Zhang2020}				&	73.2   					& 64.2     					& $\times$			\\ 
DCMN+(-L)											&	79.3						& 74.4						& $\times$   			\\ 
SG-Net(-L) \cite{zhang2019sgnet}					&	78.8  					& 72.2  						& $\times$				\\ 
CSFN(-B) \cite{Duan2021}$\bigstar$			&	68.3   					& 68.3  						& 64.0				\\ 
DUMA(-B) \cite{zhu2020duma}				&	$\times$   				& $\times$  						& 64.0				\\ 
ConceptPlug(-B)\cite{LI2021106936}$\bigstar$	&	65.3   					& 65.3   					&   65.3			\\ 
ConceptPlug(-L)$\bigstar$						&	72.6   					& 72.6   					&   69.3			\\ \hline
\hline
CTM(-B)												&	75.2   					& 68.3   					& 69.2			\\ 
CTM(-L)												&	81.5   					& 75.3   				& 72.0			\\ \hline
\hline
\end{tabular}
\end{center}
\end{table*}

\subsection{Ablation study}\label{sec:ablation}
Experiments are conducted on the RACE dataset (with the BERT-base encoder) to validate the contributions from the proposed triple-matching (TM) module and the constrative regularization. 

\textbf{On triple-matching} 
This experiment compares the performance of two different matching strategies, \emph{i.e}. the proposed TM against existing dual one. The contrastive regularization in this experiment is disabled by setting $\lambda_{CR}=0$. 

The DCMN+ model \cite{Zhang2020} is adopted as the opponent, which achieves the state-of-the-art performance. It consists of three dual-matching components: question-answer pair ($\bm{M}^{qa}$), question-passage pair ($\bm{M}^{qp}$), and answer-passage pair ($\bm{M}^{ap}$). By contrast, the proposed CTM includes three components, including $\bm{M}^{\underline{a}}$ , $\bm{M}^{\underline{p}}$, $\bm{M}^{\underline{q}}$, respectively (see Eq. (	\ref{eq:att_pqa})). Next, we carefully ablate those components by enumerating different combinations, and compare them with DCMN+ using the RACE-H dataset.

\begin{table}[!htbp]
\caption{The performance comparison of the proposed CTM against DCMN+ on the RACE-H testing set, by employing combination of different pairwise matching.}
\label{tab:tm_vs_dm}
\begin{center}\begin{tabular}{|l|c|l|c|}
\hline
\textbf{Branch}							& 	\textbf{Acc}			& 	\textbf{Branch}														& 	\textbf{Acc} 			\\ \hline
$[\bm{M}^{ap};\bm{M}^{qp}]$		&	63.8  						&  $[\bm{M}^{ap};\bm{M}^{qa}]$    	 						&	63.3						\\ \hline
$[\bm{M}^{qa};\bm{M}^{qp}]$ 		&	62.1  						&  $[\bm{M}^{ap};\bm{M}^{qa};\bm{M}^{qp}]$  			&  64.2							\\ \hline 
\hline
$\bm{M}^{\underline{a}} $ 					&	48.6						&  $\bm{M}^{\underline{q}}$    	 										&	64.1						\\ \hline
$\bm{M}^{\underline{p}} $ 					&	63.8						&  $[\bm{M}^{\underline{a}} , \bm{M}^{\underline{q}}]$ 		&	64.6						\\ \hline
$[\bm{M}^{\underline{p}} , \bm{M}^{\underline{q}} ]$ 				&	64.8						&  $[\bm{M}^{\underline{a}} , \bm{M}^{\underline{p}} ]$ 		&	52.3		\\ \hline
$[\bm{M}^{\underline{a}} , \bm{M}^{\underline{p}} , \bm{M}^{\underline{q}} ]$ 					&	65.7							& 								&						\\ \hline

\end{tabular}
\end{center}
\end{table}

Table \ref{tab:tm_vs_dm} shows the results on the proposed TM and DCMN+~\cite{Zhang2020}.
As observed, the component of $\bm{M}^{\underline{q}}$ contributes mostly in the answer selection, as it achieves the highest accuracy among all proposed components. 
This result suggests the importance of utilizing question(s) as the background context, rather than passage and/or answers, to address MCQA tasks.

On the other hand, the component of $\bm{M}^{\underline{a}}$  obtains the worst performance, which reveals the limitation of \textit{short} answers. Note that the $\bm{M}^{\underline{a}} $ is to take the answer as the background context, and estimate its correlation (using attention) between passage/question. Yet, the attention-based correlation is insignificant compared to others, mainly due to the short sequence length from answers. 

Additionally, we also notice that the combination of all three component work best with the encoder, which demonstrates a better matching outcome (65.7\%) compared to that of DCMN+ (64.2\%). The result not only indicates the necessity of utilizing all three proposed matching components, but also shows the superiority of the triple matching compared to existing dual matching.

\textbf{On contrastive regularization}
The impact of the contrastive regularization is mainly controlled by the penalty term $\lambda_{CR}$ from Eq. (\ref{eq:final}). Concretely, different settings of $\lambda_{CR}$ influence the algorithm behavior. For instance, a bigger value of $\lambda_{CR}$ will be in favor of the model via awarding the answer difference. In the extreme case of $\lambda_{CR}=0$, the model degrades to the TM module simply. Consequently, we evaluate the accuracy for CTM by setting different values to $\lambda_{CR}$ as [0, 0.5, 1, 1.5].

\begin{table}[!htbp]
\caption{Performance of the proposed CTM on RACE-H at different values of $\lambda_{CR}$.}
\label{tab:lambda_res}
\begin{center}\begin{tabular}{|c|c|c|c|c|}
\hline
$\lambda_{CR}$		& 	0 			& 	0.5 		& 1 			& 1.5 			\\ \hline
RACE-M   				&	72.9   	&  75.2   	& 73.4 		& 70.2		\\ \hline
RACE-H   				&	65.7   	&  68.3   	& 65.3 		& 62.7		\\ \hline \hline
Average    				&  67.8   	&  70.3		& 67.7 		& 64.9		 \\ \hline
\end{tabular}
\end{center}
\end{table}

With the comparison result presented in Table \ref{tab:lambda_res}, it is found out that the proposed contrastive regularization helps in enhancing the matching capability. For instance, the CTM achieves the best result when $\lambda_{CR}=0.5$, compared to that of $\lambda_{CR}=0$. Note that the latter corresponds to the simple TM module. The comparison result clearly shows the advantage of utilizing the candidate difference to improve the model answering capability.

Yet, the increase of the $\lambda_{CR}$ value results in the inferior accuracy (in particular with $\lambda_{CR}=1.5$). The reason could be the compatibility between the learned features and the final classification. With a larger $\lambda_{CR}$, the model tends to learn distinct features to separate answers, which might not be useful for selecting the correct answer.

\textbf{Analysis}
In this section, the model capability is further analyzed based on the question complexity. 
We randomly select 10\% samples (350 questions) from the testing set of RACE-H, and manually annotate them using question types of \textit{what}, \textit{which}, \textit{cloze} and \textit{other}\footnote{The ``other'' type including the rest question types, such as \textit{why}, \textit{who}, \textit{when}, \textit{where}, and \textit{how}.}. Additionally, we further tag them based on the number of sentence required to answer the question. The performance from two models is accordingly shown in Table \ref{tab:confusion}.

\begin{table}[!htbp]
\caption{Comparison of both the model performance on the RACE-H testing set, where cases are categorized by question types and the number of required sentences (\textbf{\#s}). The underlined results (\%) are from CTM, while the one within the bracket (\%) represents that of DCMN+.}
\label{tab:confusion}
\begin{center}\begin{tabular}{c|ccc}
\textbf{Type}						& 	\textbf{\#s=1} 			& 	\textbf{\#s=2} 						& \textbf{\#s$\geqslant 3$}						\\ \hline
what   								&	\underline{69.2}(68.3)	&  \underline{65.7}(63.2)  				& \underline{62.5}(58.3)		 											\\
which   								&	\underline{67.0}(65.8)  	&  \underline{67.9}(64.6)  				& \underline{66.5}(62.2)	 												\\ 
cloze   								&	\underline{70.2}(63.3)   &  \underline{70.5}(58.1)				& \underline{66.0}(55.8)		 										\\ 
other								  	&	\underline{71.3}(71.5)	&  \underline{68.7}(64.6)  				& \underline{60.5}(60.3)		 											\\ 
\end{tabular}
\end{center}
\end{table}

The result clearly indicates the superiority of the proposed algorithm when answering complex questions, such as cloze test and more sentences involved. For instance, the cloze test requires more reasoning capability as the model needs to scan the entire passage according to the given question and all candidate answers. As such, the proposed triple matching is more suitable than the conventional dual-wise strategy. Additionally, as the cloze test needs to fill in missing item(s), the textual difference from candidate answers also plays an important role. As expected, the proposed contrastive regularization helps in identifying and further highlighting those difference, thereby achieving the improvement for the question answering.

Similarly, with complex questions that need to infer from (more than) 3 sentences, the result clearly reflects an improvement from CTM compared with DCMN+. With the increasing number of required sentences, the prediction accuracy from both models has been reduced. Yet, CTM performs much stable than its counterpart, which shows its robustness of handling cases with multiple evidence sentences.
In conclusion, it can be empirically confirmed that the proposed CTM algorithm achieves comparative performance than dual-wise methods, in particular with complex question answering.

\section{Conclusion}\label{sec:conclusion}
The task of multiple choice question answering (MCQA) aims to identify a suitable answer from the background passage and question. Using the dual-based matching strategy, existing methods decouple the process into several pair-wise steps, that fail to capture the global correlation from the triple of passage, question and answer. 

In this paper, the proposed algorithm introduces a context-guided triple matching. Concretely, a triple-matching module is used to enumerate the triple and estimate a semantic matching between one component (context) with the other two. Additionally, to produce more informative features, the contrasitve regularization is further introduced to encourage the latent representation of correct answer(s) staying away from distractive ones.
Intensive experiments based on two benchmarking datasets are considered. In comparison to multiple existing approaches, the proposed algorithm produces a state-of-the-art performance by achieving higher accuracy. To our knowledge, this is the first work that explores a context-guided matching and contrasitve learning in multiple choice question answering. We will continue exploring inter/cross sentence matching as our future work.

\bibliographystyle{IEEEtran}
\bibliography{ref}
\end{document}